\documentclass[10pt,twocolumn,letterpaper]{article}

\usepackage[pagenumbers]{cvpr} 

\usepackage{multirow}
\usepackage{bbding}
\usepackage{pifont}
\usepackage{makecell}
\usepackage{rotating}
\usepackage{colortbl}
\usepackage{amssymb}

\definecolor{cvprblue}{rgb}{0.21,0.49,0.74}
\usepackage[pagebackref,breaklinks,colorlinks,allcolors=cvprblue]{hyperref}

\def\paperID{7178} 
\def\confName{CVPR}
\def\confYear{2025}

\newcommand{\NAME}{GauSTAR\xspace}

\title{\NAME: Gaussian Surface Tracking and Reconstruction}

\author{Chengwei Zheng$^1$ $\qquad$
Lixin Xue$^1$ $\qquad$
Juan Zarate$^1$ $\qquad$ 
Jie Song$^{1,2,3}$
\\
\\
$^1$ETH Zürich $\quad$
$^2$HKUST(GZ) $\quad$
$^3$HKUST\\
}

\begin{document}
\twocolumn[{
\renewcommand\twocolumn[1][]{#1}
\maketitle
\begin{center}
    \vspace{-1em}
    \centering
    \captionsetup{type=figure}
    \includegraphics[width=0.97\textwidth,trim=0 0 0 0,clip]{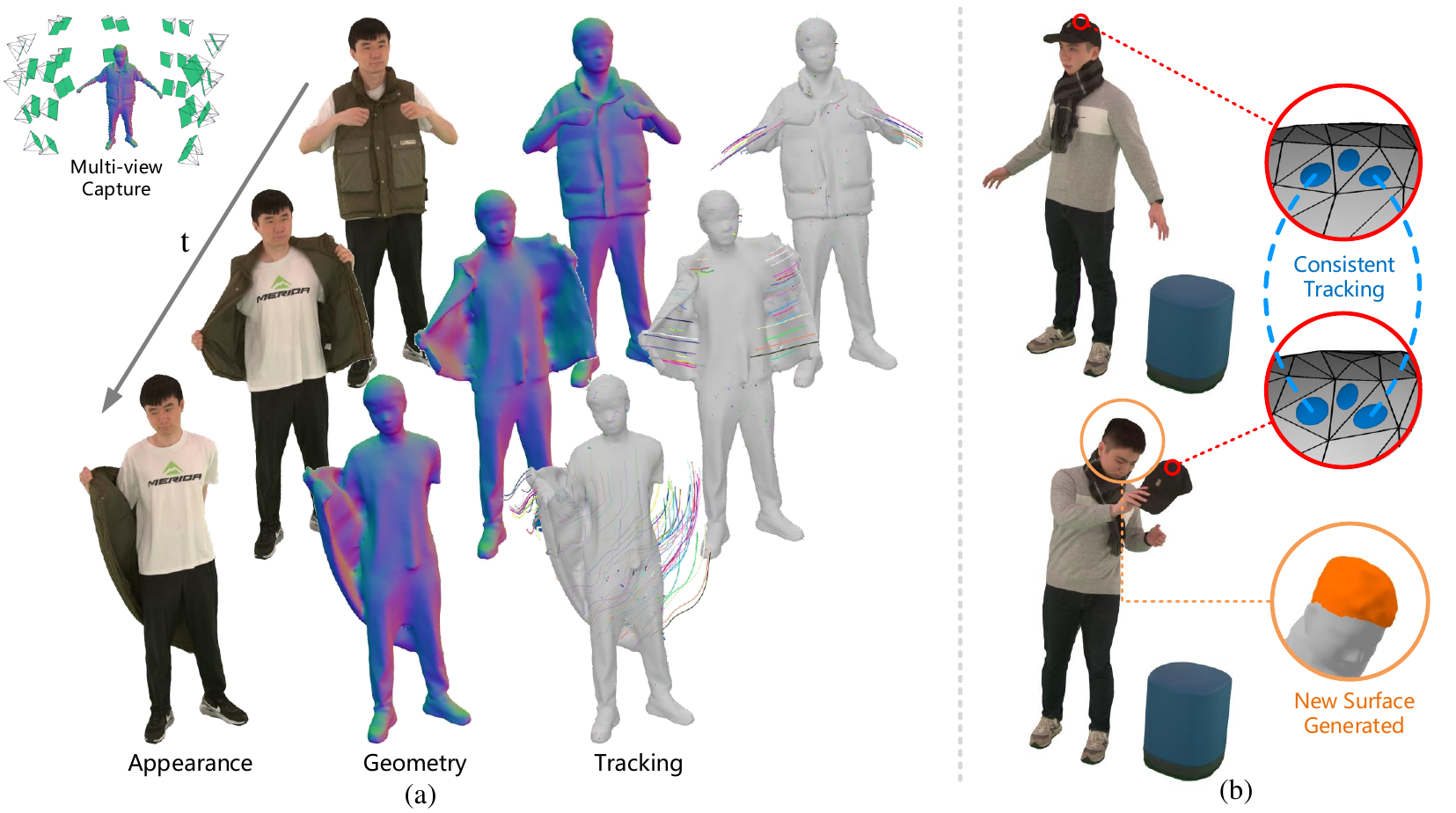}
    \vspace{-0.5em}
    \captionof{figure}{
    We propose \NAME, a novel method that (a) enables photo-realistic rendering, surface reconstruction, and 3D tracking for dynamic scenes while handling topology changes.
    (b) \NAME adapts to topology changes through two mechanisms: consistent tracking for stable surfaces (red circles) and dynamic surface generation for newly appearing geometry (orange circles).
    }
	\label{fig:teaser}
\end{center}
}]

\begin{abstract}

3D Gaussian Splatting techniques have enabled efficient photo-realistic rendering of static scenes. 
Recent works have extended these approaches to support surface reconstruction and tracking. 
However, tracking dynamic surfaces with 3D Gaussians remains challenging due to complex topology changes, such as surfaces appearing, disappearing, or splitting. 
To address these challenges, we propose \NAME, a novel method that achieves photo-realistic rendering, accurate surface reconstruction, and reliable 3D tracking for general dynamic scenes with changing topology. 
Given multi-view captures as input, \NAME binds Gaussians to mesh faces to represent dynamic objects. 
For surfaces with consistent topology, \NAME maintains the mesh topology and tracks the meshes using Gaussians. 
For regions where topology changes, \NAME adaptively unbinds Gaussians from the mesh, enabling accurate registration and generation of new surfaces based on these optimized Gaussians. 
Additionally, we introduce a surface-based scene flow method that provides robust initialization for tracking between frames. 
Experiments demonstrate that our method effectively tracks and reconstructs dynamic surfaces, enabling a range of applications.
Our project page with the code release is available at \url{https://eth-ait.github.io/GauSTAR/}.
\end{abstract}
\section{Introduction}
\label{sec:intro}

In the realm of dynamic scene representations, we seek methods capable of delivering photorealistic renderings from arbitrary viewpoints, as well as surface reconstructions that adapt to changing topologies. 
Scenarios involving human or robotic interactions with objects require dynamic adaptation to surfaces that split, merge, or deform. 
Furthermore, downstream applications such as visual effects and markerless motion capture benefit significantly from the ability to track persistent regions over time without relying on templates. 
Therefore, methods must efficiently handle these topology changes to ensure high-quality renderings and accurate reconstruction while also maintaining consistent tracking of existing surfaces.

Classical methods primarily rely on meshes and texture maps, which provide reasonable appearances but heavily depend on mesh resolution. They often fail to render fine details and view-dependent effects accurately. Although these mesh representations allow for some level of tracking, they struggle to handle significant topology changes, necessitating new keyframes to accommodate major transformations. 
The advent of Neural Radiance Fields (NeRF)~\cite{mildenhall2020nerf} brought significant improvements in appearance and novel view synthesis for both static~\cite{barron2022mip, zhang2020nerf++} and dynamic scenes~\cite{park2021nerfies, icsik2023humanrf}.
While surfaces can be derived from implicit Signed Distance Functions (SDF) using Marching Cubes~\cite{yu2022monosdf, wang2021neus}, they lack consistent tracking unless underlying templates are used. 
Recently, 3D Gaussian Splatting (3DGS)~\cite{kerbl20233d} has emerged with explicit texture representation, rivaling NeRF in appearance while achieving more efficient renderings. Its explicit representation facilitates tracking, and several techniques have been developed for this purpose~\cite{luiten2024dynamic, zheng2024physavatar}. However, accurate dynamic surface reconstruction remains a challenge, and balancing the tracking of existing surfaces with the introduction of new ones proves difficult. 

To address these challenges, we propose \NAME, a method capable of reconstructing photorealistic appearances and accurate surface geometries with consistent tracking as topology changes. 
\NAME leverages multi-view capture and combines meshes with bound Gaussians to create Gaussian Surfaces. 
These Gaussians move along with the mesh faces to represent objects that move and deform. 
When new surfaces become visible, new Gaussians are generated, and the mesh topology updates. 
The adaptable mesh provides a time-consistent and accurate geometry, while the Gaussians bring a photorealistic appearance.

This problem is challenging because there is always a trade-off. 
Methods that allow easier tracking via fixed topologies or templates~\cite{liu2024dynamic, zheng2024physavatar} tend to degrade the quality of the appearance and geometry under new poses or deformations.
Conversely, methods that overfit static scenes~\cite{guedon2024sugar, huang20242d, dai2024high} lack temporal consistency or miss new frame details.
\NAME addresses this trade-off by tracking as many surfaces over time as possible while remaining flexible to enable new faces and Gaussians to appear where the topology changes. 
We adapt the preceding frame by deforming the mesh and optimizing Gaussian parameters. 
For topology-changing surfaces such as newly emerging ones, \NAME first unbinds the Gaussians in these regions, allowing them to move beyond the mesh faces. New Gaussian Surfaces are then generated based on the unbound Gaussians, enabling accurate reconstruction of the new surfaces. 
Additionally, we propose a surface-based scene flow method that back-projects 2D optical flow into 3D space using depth images. This method provides an initialization for frame-by-frame tracking to robustly manage large 3D or fast motions.

Our contributions are as follows.
\begin{itemize}
\item A new framework for tracking and reconstructing dynamic scenes combining 3D Gaussians and meshes, effectively managing changes in topology. 
\item A method for Gaussian unbinding and surface re-meshing that allows the generation of new surfaces as topologies evolve. 
\item A method for handling large or fast deformation of surfaces between frames via scene flow warping.
\end{itemize}

As demonstrated in our experiments, \NAME matches or surpasses SOTA methods in appearance metrics, thanks to the performance of Gaussian Surfaces. 
This makes it a strong representation for high-quality 3D rendering applications such as VR/XR and telepresence. 
Simultaneously, \NAME provides a high-resolution, explicit 3D representation with robust tracking capabilities, as illustrated by our AprilTag-based experiments. 
We expect these tracked meshes will facilitate numerous tasks beyond rendering, benefiting fields such as computer vision, computer graphics, robotics, biomechanics, spatial audio, and more.

\section{Related Work}
\label{sec:rela_work}

\subsection{3D Neural Representations}
The reconstruction of general scenes has been a long-standing problem in computer vision. Traditional methods relied on triangle meshes and texture maps~\cite{starck2007surface, collet2015high, zheng2021dtexfusion, lin2022occlusionfusion, zheng2022self}. While these explicit representations enabled efficient rendering and intuitive geometry editing, they struggled with view-dependent effects and fine surface details.
Recent neural representations such as NeRF~\cite{mildenhall2020nerf} and 3DGS~\cite{kerbl20233d} have significantly advanced the static reconstruction field with coordinate-based networks and explicit 3D Gaussians. 
Both NeRF and 3DGS can achieve photo-realistic rendering, while the surfaces cannot be extracted accurately. 
NeRF-based methods address this by introducing SDF fields~\cite{yu2022monosdf, wang2021neus, wang2023neus2} and using Marching Cubes~\cite{lorensen1987marching} to generate surfaces.
Gaussian-based methods propose to use surface-alignment regularization terms~\cite{guedon2024sugar, dai2024high, fan2024trim, choi2024meshgs} during training, followed by mesh extraction.
As 3D Gaussians are discrete presentations rather than continuous representations like NeRF, Poisson reconstruction~\cite{guedon2024sugar, dai2024high} and SDF fusion~\cite{wolf2024gs2mesh, huang20242d, chen2024vcr} are also employed for mesh extraction, along with Marching Cubes~\cite{li2024mvg, lin2024direct, Yu2024GOF}. 
To further improve the quality, Gaussian Surfels~\cite{dai2024high} propose a novel point-based representation and a self-supervised normal-depth consistency regularizer. 
2D Gaussian Splatting~\cite{huang20242d} uses 2D Gaussians that are tightly aligned to surfaces.
NeuSG~\cite{chen2023neusg} jointly optimizes implicit surface reconstruction with 3D Gaussian Splatting.
While these methods perform well for static scenes, applying them to achieve consistent dynamic reconstruction is far from straightforward.

\subsection{4D Dynamic Representations}

To handle dynamic scenes, several NeRF-based methods~\cite{park2021hypernerf, zheng2023editablenerf, fridovich2023k, wang2023neus2} incorporate time-dependent variables into the reconstruction model to represent movements and deformations. 
For instance, HumanRF~\cite{icsik2023humanrf} reconstructs radiance fields using 4D feature grids over temporal segments. While these approaches achieve photo-realistic rendering of general dynamic scenes, they lack consistent tracking capabilities, limiting their practical applications.

For 3DGS-based dynamic surface reconstruction, deformation modules~\cite{liu2024dynamic, cai2024dynasurfgs, zhang2024dynamic} are also introduced to deform Gaussians and reconstruct meshes from monocular inputs. 
MaGS~\cite{ma2024reconstructing} constrains 3D Gaussians to hover on the mesh surface, creating a mutual-adsorbed mesh-Gaussian representation.
Space-time 2D Gaussian Splatting~\cite{wang2024space} leverages 2D Gaussians to reconstruct dynamic scenes and extract surfaces from them. 
However, these methods extract meshes independently for each frame, limiting their ability to generate face correspondences across frames.

\subsection{Tracking}

Tracking methods aim to estimate the motion trajectories of surface points. 
2D tracking methods~\cite{doersch2022tap, harley2022particle, doersch2023tapir, karaev2023cotracker} take video inputs and track pixels across frames.
OmniMotion~\cite{wang2023tracking} improves pixel-wise tracking by introducing a 3D canonical volume and a set of bijections.
Shape of Motion~\cite{wang2024shape} represents scene motion using a set of motion bases, providing a globally consistent representation of dynamic scenes. 
With the use of 3D Gaussians as an explicit representation, new possibilities emerge for more efficient 3D tracking.
Dynamic 3D Gaussians~\cite{luiten2024dynamic} tracks Gaussians by directly optimizing their positions with multi-view inputs. 
PhysAvatar~\cite{zheng2024physavatar} tracks time-consistent meshes and models human clothes via physics-based simulation and rendering. 
To improve performance, traditional non-rigid fusion techniques~\cite{jiang2024hifi4g}, optical flow methods~\cite{zhu2024motiongs, chu2024dreamscene4d, gao2024gaussianflow}, and multi-head deformation decoders~\cite{wu20244d} are also employed.
Additionally, some Gaussian-based reconstruction methods~\cite{li2024spacetime, duan20244d} deform sets of Gaussians to represent dynamic scenes; however, their use of temporary Gaussians restricts them to short-range tracking.
While these methods demonstrate the ability to track Gaussians or meshes with fixed topology, tracking surfaces as topologies evolve remains an open challenge.
\section{Method}
\label{sec:method}

\begin{figure*}[t]
\includegraphics[width=\linewidth, trim=0 20 0 0,clip]{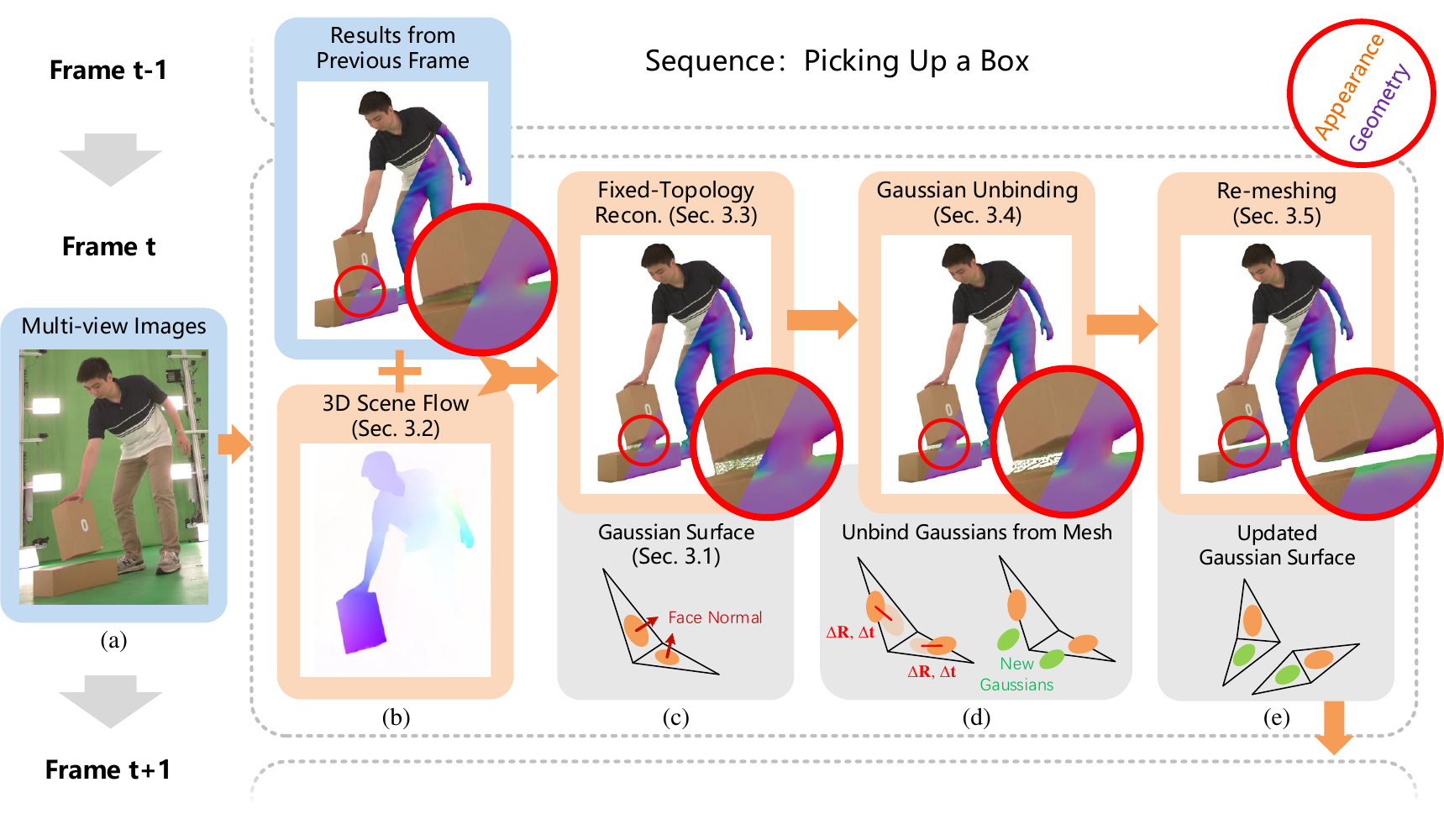}
\caption{Taking multi-view captures as input, \NAME tracks and reconstructs dynamic objects frame by frame. For each frame, \NAME first warps the previous frame's result using scene flow (\cref{sec:scene_flow_warping}). It then reconstructs Gaussian Surfaces (\cref{sec:gaussian_surface}) by fixed-topology reconstruction (\cref{sec:basic_tracking}). To handle topology-changing surfaces, \NAME detects topology changes, unbinds Gaussians on these surfaces, and adds new Gaussians as needed (\cref{sec:Gaussian_unbinding}). Finally, the Gaussian Surfaces are updated through re-meshing (\cref{sec:surface_remeshing}).}
\label{fig:pipeline}
\end{figure*}

Our system takes multi-view RGB-D videos as input. 
We aim to achieve consistent reconstruction and tracking even when surfaces undergo topology changes. 
To represent dynamic objects, we introduce Gaussian Surfaces - meshes with Gaussians attached to their faces - which enable both accurate geometry reconstruction and photo-realistic rendering (\cref{sec:gaussian_surface} and \cref{fig:pipeline} (c)). 
For each frame, we first initialize surface positions through scene flow warping from the previous frame (\cref{sec:scene_flow_warping} and \cref{fig:pipeline} (b)).
We then optimize the Gaussian Surfaces with the topology from the previous frame using multi-view constraints (\cref{sec:basic_tracking} and \cref{fig:pipeline} (c)). 
For regions experiencing topology changes, which are detected through our novel Gaussian unbinding weights, we allow Gaussians to detach from the original mesh faces and optimize their positions independently (\cref{sec:Gaussian_unbinding} and \cref{fig:pipeline} (d)). 
Lastly, we perform re-meshing to update the topology-changing geometry, and ensure our representation remains consistent in other regions (\cref{sec:surface_remeshing} and \cref{fig:pipeline} (e)).

\subsection{Gaussian Surface Representation}
\label{sec:gaussian_surface}

Our Gaussian Surface representation, shown in (c) of \cref{fig:pipeline}, augments traditional meshes with $N$ Gaussians per triangular face~\cite{guedon2024sugar, zheng2024physavatar}. 
Following the formulation from 3DGS~\cite{kerbl20233d}, a 3D Gaussian can be represented as:
\begin{equation}
  G(\mathbf{x}) = \sigma(\alpha)  \cdot \mathbf{exp} \left ( -\frac{1}{2}(\mathbf{x}-\mathbf{p})^\top \Sigma^{-1}(\mathbf{x}-\mathbf{p}) \right ),
  \label{eq:gs_exp}
\end{equation}
\begin{equation}
  \Sigma = \mathbf{R} \mathbf{S} \mathbf{S}^\top \mathbf{R}^\top.
  \label{eq:gs_rs}
\end{equation}
Here a Gaussian $G$ is defined by its opacity $\alpha$, center position $\mathbf{p}$, scales $\mathbf{S}$, rotation $\mathbf{R}$, and appearance color (represented by spherical harmonics). Moreover, $\sigma()$ is the standard sigmoid function, and $\Sigma$ denotes the covariance matrix.

To construct Gaussian Surfaces, we uniformly distribute $N$ Gaussians on each triangular face. Each Gaussian center $\mathbf{p}$ is computed from the face vertices $\mathbf{v}_1, \mathbf{v}_2,\mathbf{v}_3 $ using its predefined barycentric coordinate $(b_1, b_2, b_3)$.
\begin{equation}
  \mathbf{p} = b_1 \mathbf{v}_1 + b_2 \mathbf{v}_2 + b_3 \mathbf{v}_3.
  \label{eq:gs_bary}
\end{equation}
To ensure Gaussians remain aligned with the mesh surface, we constrain their orientation and thickness: the z-axis $\mathbf{R}_z$ is set to the face normal $\mathbf{n}(f)$, while the z-scale $\mathbf{S}_z$ is set to a small predefined value $\delta$.
The remaining Gaussian parameters are jointly optimized following the volumetric rendering approach of 3DGS~\cite{kerbl20233d}.

\subsection{Scene Flow Warping}
\label{sec:scene_flow_warping}

Dynamic scenes often exhibit large or rapid deformations between frames, making tracking and optimization challenging. To address this, we estimate 3D scene flow by re-projecting 2D optical flow using depth information, providing robust initialization for each frame's reconstruction.

Given adjacent frames $t$ and $t+1$, we compute scene flow in four steps. First, we project each vertex from frame $t$ into all visible input views. Second, we compute the corresponding pixel positions in frame $t+1$ using optical flow ~\cite{teed2020raft}. Third, we re-project these 2D positions in $t+1$ back to 3D using the respective depth images. Finally, we aggregate the 3D movements between frames across all views to obtain the scene flow for each vertex.

To enhance robustness, we filter out unreliable flows by performing bi-directional optical flow consistency checks between frames $t$ and $t+1$, along with depth discontinuity detection on the depth images.
We further refine the scene flow through surface-aware smoothing:
\begin{equation}
\mathcal{F}'(v) = \frac{1}{\left| \mathbf{N}(v) \right|}\sum_{u \in \mathbf{N}(v)} w(u, v) \mathcal{F}(u),
\label{eq:scene_flow_smooth}
\end{equation}
where $\mathcal{F}(v)$ and $\mathcal{F}'(v)$ are the scene flow before and after smoothing, $\mathbf{N}(v)$ represents mesh-connected neighbors of vertex $v$, and $w(u,v)$ weights neighbors by their distance. The final vertex positions are updated as $v + \mathcal{F}'(v)$.

\subsection{Fixed-Topology Surface Reconstruction}
\label{sec:basic_tracking}

We first reconstruct Gaussian Surfaces assuming fixed topology, establishing a baseline reconstruction that will later be refined to handle topology changes.
Given multi-view RGB-D inputs and a mesh initialized through scene flow warping, we optimize vertex positions and Gaussian parameters using RGB, depth, and mask supervision:
\begin{equation}
  \mathcal{L}_{\text{rgb}} = \left\| \hat{C}_i - C_i \right\|_1 + \lambda_{\text{SSIM}} \mathcal{L}_{\text{SSIM}}(\hat{C}_i, C_i),
  \label{eq:rgb_loss}
\end{equation}
\begin{equation}
  \mathcal{L}_{\text{depth}} = \left\| \hat{D}_i - D_i \right\|_1, \hspace{0.5em} \mathcal{L}_{\text{mask}} = \left\| \hat{M}_i - M_i \right\|_1.
  \label{eq:depth_loss}
\end{equation}
Here $\hat{C}_i, \hat{D}_i, \hat{M}_i$ represent predicted color, depth, and mask images from Gaussian rendering, while $C_i, D_i, M_i$ are the input images. 
As Gaussian positions are determined by vertex positions in \cref{eq:gs_bary}, optimizing these losses effectively updates the underlying mesh.

We further introduce three regularization terms: normal smoothing to ensure surface continuity, area preservation to maintain local geometry, and color consistency to promote temporal coherence:
\begin{equation}
  \mathcal{L}_{\text{smooth}} = \frac{1}{\left| \mathbf{F} \right|} \sum_{f_i \in \mathbf{F}} \sum_{f_j \in \mathbf{N}(f_i)} \left  (1-\mathbf{n}(f_i)\cdot \mathbf{n}(f_j) \right ),
  \label{eq:smooth_loss}
\end{equation}
\begin{equation}
  \mathcal{L}_{\text{area}} = \frac{1}{\left| \mathbf{F} \right|} \sum_{f \in \mathbf{F}} \left\| \mathbf{Area}(f;t) - \mathbf{Area}(f;0) \right\|_1,
  \label{eq:area_loss}
\end{equation}
\begin{equation}
  \mathcal{L}_{\text{SH}} = \frac{1}{\left| \mathbf{G} \right|} \sum_{g \in \mathbf{G}} \left\| \mathbf{SH}(g;t) - \mathbf{SH}(g;t\!-\!1) \right\|^2_2,
  \label{eq:consis_loss}
\end{equation}
where $\mathbf{F}$ is the set of triangular faces, $\mathbf{N}(f_i)$ denotes neighboring faces of $f_i$, and $\mathbf{n}(f)$ computes the face normal. $\mathbf{Area}(f;t)$ measures face area at frame $t$, penalizing deviation from initial areas $\mathbf{Area}(f;0)$. $\mathbf{SH}(g;t)$ represents the spherical harmonics parameters of Gaussian $g$ at frame $t$, where $\mathbf{G}$ denotes the set of Gaussians. Additionally, we constrain each Gaussian's scale by its face's edge length and enforce a minimum opacity to ensure surface opacity.

\subsection{Adaptive Gaussian Unbinding}
\label{sec:Gaussian_unbinding}

While fixed-topology surface reconstruction refines meshes based on multi-view input, it cannot handle emerging surfaces or topology changes. We address this problem by allowing Gaussians to detach from the mesh faces in regions where topology changes are detected.

As illustrated in \cref{fig:pipeline} (d),  we introduce additional transformation parameters for each Gaussian: a rotation $\Delta \mathbf{R}$ applied to the Gaussian orientation $\mathbf{R}$, and a translation $\Delta \mathbf{t}$ added to its center position $\mathbf{p}$. 
To optimize these transformations, we extend the fixed-topology reconstruction process to jointly optimize both the original parameters and these additional transformations. This allows Gaussians to move independently from their underlying mesh faces when necessary.

\begin{figure}[t]
\includegraphics[width=\linewidth, trim=0 0 0 0,clip]{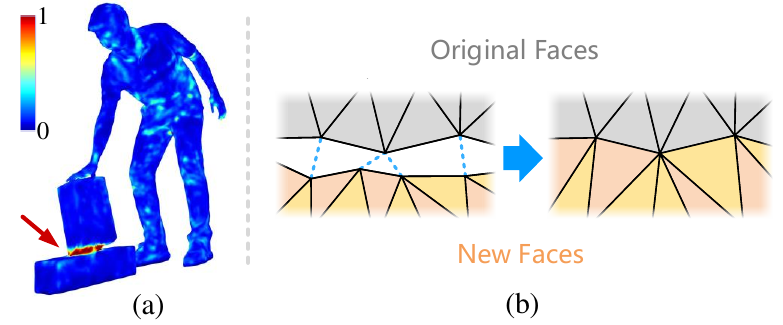}
\caption{Details of the mesh update process. (a) Visualization of unbinding weights defined in \cref{eq:unbind_weight}, where red indicates high weights in topology-changing regions. 
(b) Mesh connection process between original and new surfaces, with blue dotted lines showing vertex correspondences.}
\label{fig:method_explain}
\end{figure}

To identify which Gaussians should be unbound from the mesh, we develop a weighting scheme based on geometric and photometric cues.
Inspired by the adaptive density control in 3DGS~\cite{kerbl20233d}, we observe that topology changes typically manifest as large positional gradients and high reconstruction errors, as shown in the top part of \cref{fig:pipeline} (c). 
Based on this, we define an \emph{unbinding weight} for each face $f$:
\begin{equation}
  \mathcal{W}(f) = \mathcal{G}_{\text{pos}}(f) + \lambda_{\text{rgb}} \mathcal{L}_{\text{rgb}}(f) + \lambda_{\text{depth}} \mathcal{L}_{\text{depth}}(f).
  \label{eq:unbind_weight}
\end{equation}
Here unbinding weight $\mathcal{W}$ measures the likelihood of topology changes for each face $f$. It combines positional gradients $\mathcal{G}_{\text{pos}}$~\cite{kerbl20233d} and reconstruction errors from fixed-topology optimization. 
We cap $\mathcal{W}$ at 1 and visualize an example in \cref{fig:method_explain} (a).

Based on these unbinding weights, we introduce a regularization term to control the extent of transformations for Gaussians:
\begin{equation}
  \mathcal{L}_{\text{unb}} (g) =
  \left(1 \! - \! \mathcal{W} (f_g) \right) \left( \left\| \Delta \mathbf{R}(g) \! - \! \mathbf{I} \right\|_1 + \lambda_{\text{t}} \left\| \Delta \mathbf{t}(g) \right\|_1 \right).
  \label{eq:unbind_loss}
\end{equation}
This loss regulates the transformation of each Gaussian $g$ on face $f_g$. When unbinding weight $\mathcal{W}$ is high, indicating likely topology changes, the loss term weight becomes small, allowing larger transformations and effectively unbinding the Gaussian from its face.

In regions experiencing significant topology changes, we not only allow existing Gaussians to detach but also introduce new ones. As shown in~\cref{fig:pipeline} (d),  for faces where unbinding weights exceed a threshold, we duplicate their associated Gaussians. 
These newly introduced Gaussians are also unbound and governed by the unbinding regularization loss in~\cref{eq:unbind_loss}.

\subsection{Surface Re-meshing}
\label{sec:surface_remeshing}

\begin{figure*}[t]
\includegraphics[width=\linewidth, trim=0 0 0 0,clip]{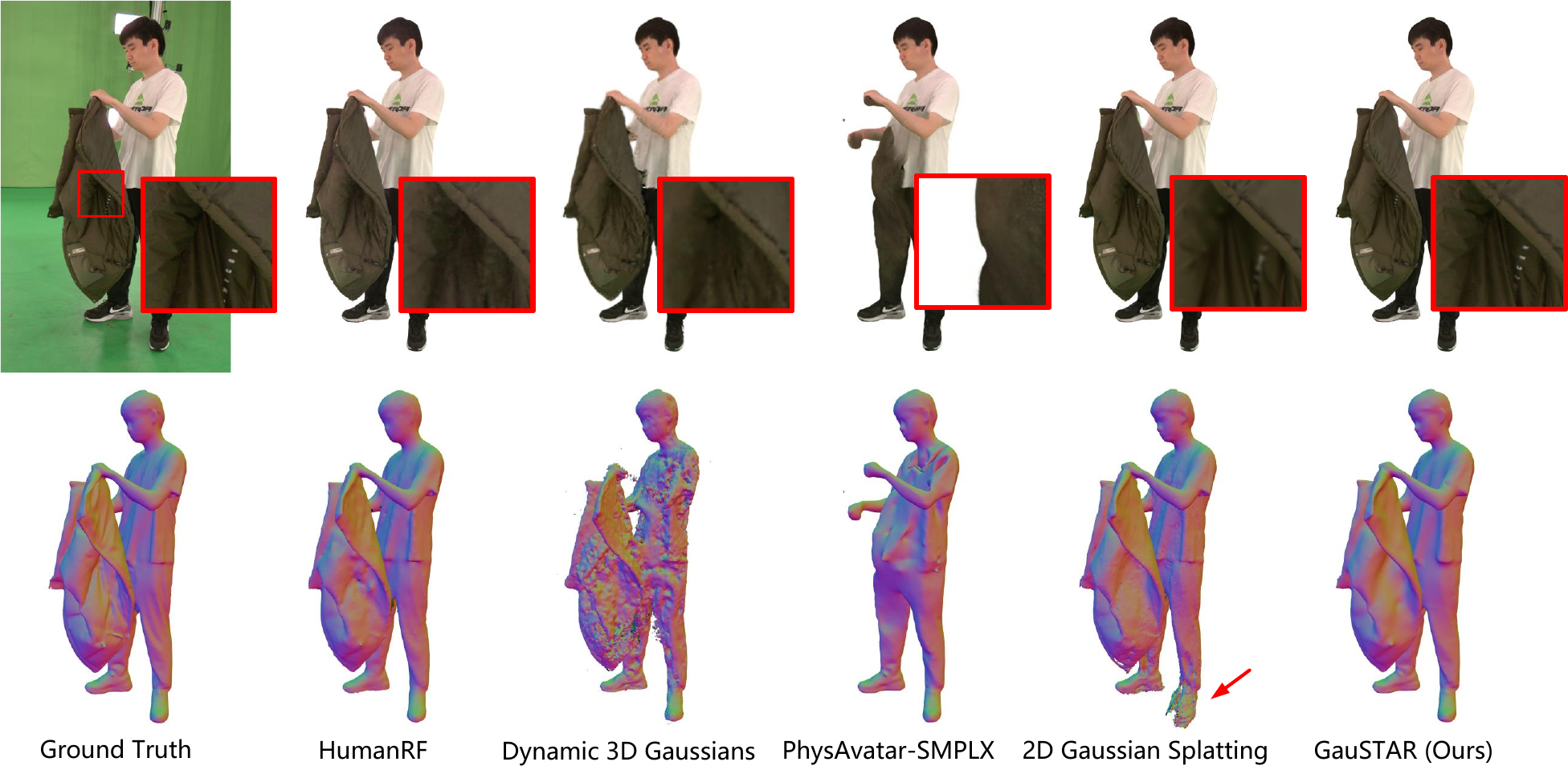}
\caption{Comparisons of appearance and geometry reconstruction. Dynamic 3D Gaussians~\cite{luiten2024dynamic} and PhysAvatar~\cite{zheng2024physavatar} yield suboptimal reconstruction results. HumanRF~\cite{icsik2023humanrf} and 2DGS~\cite{huang20242d}, lacking tracking capabilities, struggle under heavy occlusion. In contrast, \NAME provides high-quality reconstruction while supporting tracking. Additional comparisons are provided in our supplementary materials.}
\label{fig:comparison}
\end{figure*}

After Gaussian unbinding, we face two key technical challenges: identifying which mesh regions to update and ensuring smooth transitions at region boundaries. We address these by first localizing topology changes through unbinding weights, then selectively reconstructing and integrating new surfaces while preserving the original mesh structure elsewhere.

To generate new surfaces from unbound Gaussians, we first render depth images from multiple views, including capture views and uniformly sampled views on a surrounding sphere. We then employ a TSDF fusion method~\cite{newcombe2011kinectfusion} to reconstruct new meshes from these depth images.

After obtaining new meshes, we identify regions on the original mesh for replacement where unbinding weights exceed a predefined threshold. Since the unbound Gaussians from these regions have moved to new surface locations during optimization, we construct a voxel volume to locate these Gaussian positions. Within this volume, we identify corresponding faces from the new meshes to replace the original faces, and then connect them with the remaining original mesh.

To connect old and new meshes, we establish vertex correspondences along their boundaries, as shown in \cref{fig:method_explain} (b). 
Each boundary vertex from one mesh is matched to its closest vertices on the other mesh boundary, 
allowing for one-to-many or many-to-one correspondences. 
We merge matched vertices into a single vertex, and then perform post-processing steps, including edge flipping and hole filling~\cite{sharp2020you}, to refine the boundary region.

At the boundary of topology-changing surfaces, the unbinding weights gradually change from $1$ (fully unbound) to $0$ (fully bound), as illustrated in \cref{fig:method_explain} (a). Accordingly, the constraints on Gaussian transformations $\Delta \mathbf{R}, \Delta \mathbf{t}$ in~\cref{eq:unbind_loss} vary smoothly, ensuring continuity between original and new surfaces at their connection. 
After re-meshing, we perform an additional round of fixed-topology reconstruction described in~\cref{sec:basic_tracking} to fine-tune the updated Gaussian Surfaces.

\section{Experiments}
\label{sec:experiment}

\begin{table*}
  \centering
  \begin{tabular}{@{}l|ccc|cc|cc@{}}
    \toprule
    \multirow{2}*{Method} & \multicolumn{3}{c|}{Appearance} & \multicolumn{2}{c|}{Geometry} & \multicolumn{2}{c}{Tracking} \\
    ~ & PSNR $\uparrow$ & SSIM $\uparrow$ & LPIPS $\downarrow$ & CD $\downarrow$ & F-Score $\uparrow$ & 3D ATE $\downarrow$ & 2D ATE $\downarrow$ \\
    \midrule
    HumanRF~\cite{icsik2023humanrf} & \cellcolor{orange!30}{30.59} & \cellcolor{orange!30}{0.947} & \cellcolor{yellow!30}{0.128} & \cellcolor{orange!30}{0.284} & \cellcolor{orange!30}{0.968} & - & - \\
    Dynamic 3D Gaussians~\cite{luiten2024dynamic} & 27.61 & 0.905 & 0.214 & \hspace{0.66em}1.113 $^\dag$ & \hspace{0.66em}0.733 $^\dag$ & \cellcolor{yellow!30}{3.15} & \cellcolor{yellow!30}{13.84} \\
    PhysAvatar-general~\cite{zheng2024physavatar} & 22.69 & 0.893 & 0.216 & 1.372 & 0.793 & 12.94 & 56.95 \\
    PhysAvatar-SMPLX~\cite{zheng2024physavatar} & 24.50 & 0.908 & 0.193 & 0.625 & 0.837 & 8.98 & 39.61 \\
    2D Gaussian Splatting~\cite{huang20242d} & \cellcolor{yellow!30}{30.17} & 0.938 & 0.155 & 0.699 & 0.946 & - & - \\
    \NAME w/o IR input & 30.05 & \cellcolor{yellow!30}{0.946} & \cellcolor{orange!30}{0.110} & \cellcolor{yellow!30}{0.335} & \cellcolor{yellow!30}{0.960} & \cellcolor{orange!30}{0.671} & \cellcolor{orange!30}{3.02} \\
    \NAME (Ours) & \cellcolor{red!30}{\textbf{31.87}} & \cellcolor{red!30}{\textbf{0.952}} & \cellcolor{red!30}{\textbf{0.102}} & \cellcolor{red!30}{\textbf{0.237}} & \cellcolor{red!30}{\textbf{0.980}} & \cellcolor{red!30}{\textbf{0.452}} & \cellcolor{red!30}{\textbf{2.03}} \\
    \bottomrule
  \end{tabular}
  \caption{Quantitative comparisons on appearance, geometry, and tracking. The \colorbox{red!30}{best}, \colorbox{orange!30}{second-best}, and \colorbox{yellow!30}{third-best} results are highlighted. Our method achieves the best performance on reconstruction and tracking. CD and 3D ATE are reported in cm. $\dag$: Dynamic 3D Gaussians~\cite{luiten2024dynamic} doesn't provide surface reconstruction and we extract per-frame meshes using TSDF fusion~\cite{huang20242d}.}
  \label{tab:comparison}
\end{table*}

\subsection{Implement Details}
We use a capture studio with 52 RGB cameras and 52 IR cameras for capturing. Sequences are captured at a resolution of $3004 \times 4092$ and 30 fps. Unstructured IR laser lights are used, allowing us to generate depth images from the IR captures. Specifically, we generate raw point clouds from the IR images, refine them using the method in~\cite{collet2015high}, and project them onto the camera views. We attach $N=6$ Gaussians per face. The initial mesh for the first frame can be obtained using any multi-view reconstruction method and we use \cite{collet2015high}. Additional implementation details, along with code release information for public datasets, are available in our supplementary materials.

\begin{table*}
  \centering
  \begin{tabular}{@{}l|ccc|cc|cc@{}}
    \toprule
    \multirow{2}*{Method} & \multicolumn{3}{c|}{Appearance} & \multicolumn{2}{c|}{Geometry} & \multicolumn{2}{c}{Tracking} \\
    ~ & PSNR $\uparrow$ & SSIM $\uparrow$ & LPIPS $\downarrow$ & CD $\downarrow$ & F-Score $\uparrow$ & 3D ATE $\downarrow$ & 2D ATE $\downarrow$ \\
    \midrule
    \NAME w/o unbinding & 29.30 & 0.940 & 0.132 & 0.411 & 0.938 & 2.85 & 12.78 \\
    \NAME w/o re-meshing & 29.77 & 0.943 & 0.129 & 0.418 & 0.936 & 2.08 & 9.16 \\
    \NAME w/o scene flow & 29.92 & 0.943 & 0.127 & 0.433 & 0.916 & 6.56 & 29.96 \\
    \NAME (Ours final) & \textbf{31.87} & \textbf{0.952} & \textbf{0.102} & \textbf{0.237} & \textbf{0.980} & \textbf{0.45} & \textbf{2.03} \\
    \bottomrule
  \end{tabular}
  \caption{Ablations. Quantitative evaluation of each key component: Gaussian unbinding, surface re-meshing, and scene flow warping. Results demonstrate that each component is integral to the final reconstruction and tracking quality.}
  \label{tab:ablation}
\end{table*}

\subsection{Comparisons}
We compare our method with SOTA multi-view reconstruction methods HumanRF~\cite{icsik2023humanrf}, Dynamic 3D Gaussianss~\cite{luiten2024dynamic}, PhysAvatar~\cite{zheng2024physavatar}, and 2D Gaussian Splatting (2DGS)~\cite{huang20242d}, as shown in \cref{tab:comparison} and \cref{fig:comparison}. 1) HumanRF is a NeRF-based method for dynamic scene reconstruction. Its implicit representations do not provide correspondence or tracking information between frames. Since its mesh extraction code is not publicly available, we implemented it based on the paper. 2) Dynamic 3D Gaussians extend 3DGS~\cite{kerbl20233d} to handle dynamic scenes, enabling Gaussian tracking but without producing a surface mesh. Extracting meshes from its Gaussians results in noisy outputs. 3) PhysAvatar tracks an initial mesh using Gaussians in its first stage, with subsequent stages focusing on clothing reconstruction through simulation. We compare two versions of its first stage: one for general objects and another tailored for human subjects using SMPL-X~\cite{pavlakos2019expressive}. PhysAvatar is limited in handling topology changes and struggles with large deformations.  4) 2DGS uses flat Gaussians for surface reconstruction, which we apply independently to each frame in our experiments. Without incorporating temporal information, 2DGS converges to artifacts in several frames.

To ensure a fair comparison, we implemented a new version of our method without IR depth inputs, using rendered depth images from HumanRF~\cite{icsik2023humanrf} as the depth inputs. Additional experiment details and results are provided in our supplementary materials.

\textbf{Appearance.}
We evaluate the quality of appearance reconstruction through novel view synthesis. The experiment includes 4 sequences (totaling 850 frames) and 5 test views. We report PSNR, SSIM, and LPIPS metrics in \cref{tab:comparison} and present qualitative comparisons in \cref{fig:comparison}. Our method achieves superior appearance quality, particularly on surfaces with significant occlusion, benefiting from the ability to track prior surface appearance.

\textbf{Geometry.}
For geometry comparison, we use an RGB-D multi-view reconstruction method~\cite{collet2015high} to obtain the ground truth meshes. Chamfer Distance (CD, in cm) and F-Score~\cite{tatarchenko2019single} are reported in \cref{tab:comparison}, with qualitative comparisons shown in \cref{fig:comparison}. Since Dynamic 3D Gaussians~\cite{luiten2024dynamic} does not provide surface reconstruction, we extract per-frame meshes using TSDF fusion~\cite{huang20242d}, following a similar approach to our new face generation. Our method yields high-quality geometry while maintaining surface tracking.

\begin{figure}[t]
\includegraphics[width=\linewidth, trim=0 0 0 0,clip]{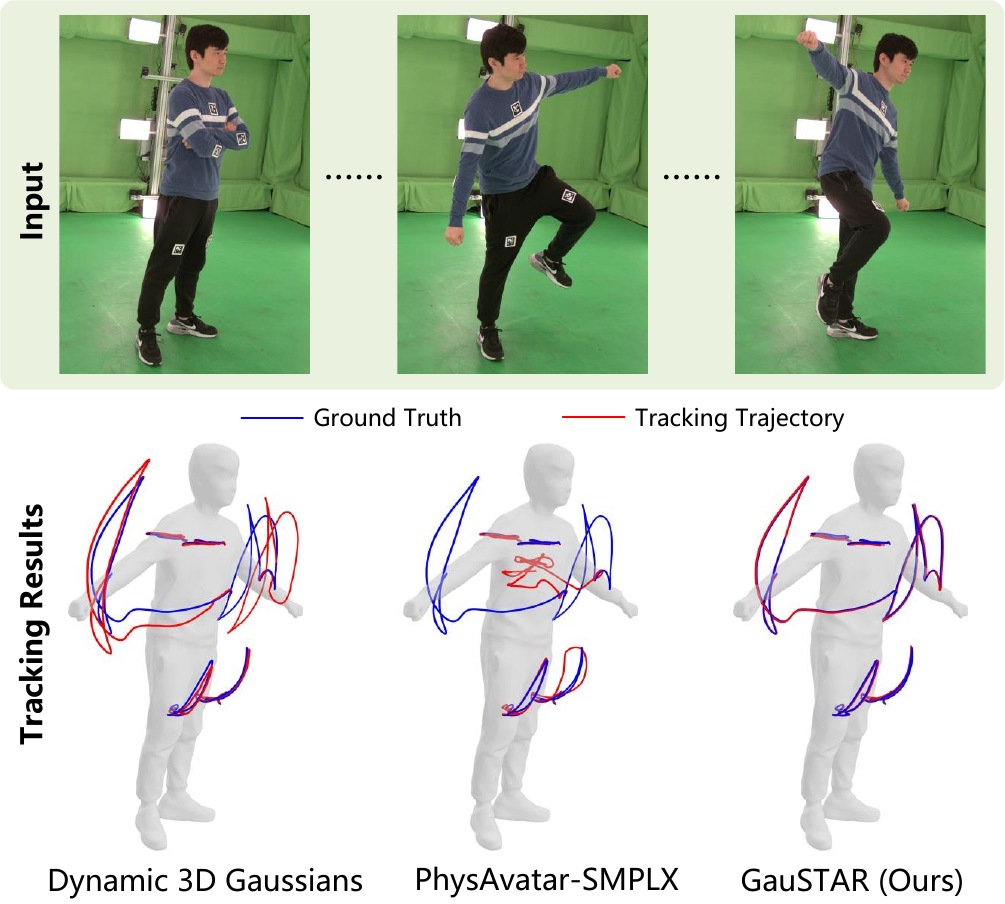}
\caption{
Tracking comparisons using AprilTags. \NAME achieves more accurate tracking results, with predicted (red) and ground truth (blue) trajectories of tag centers shown.
}
\label{fig:tracking_comp}
\end{figure}

\begin{figure*}[ht]
\includegraphics[width=\linewidth, trim=0 0 0 0,clip]{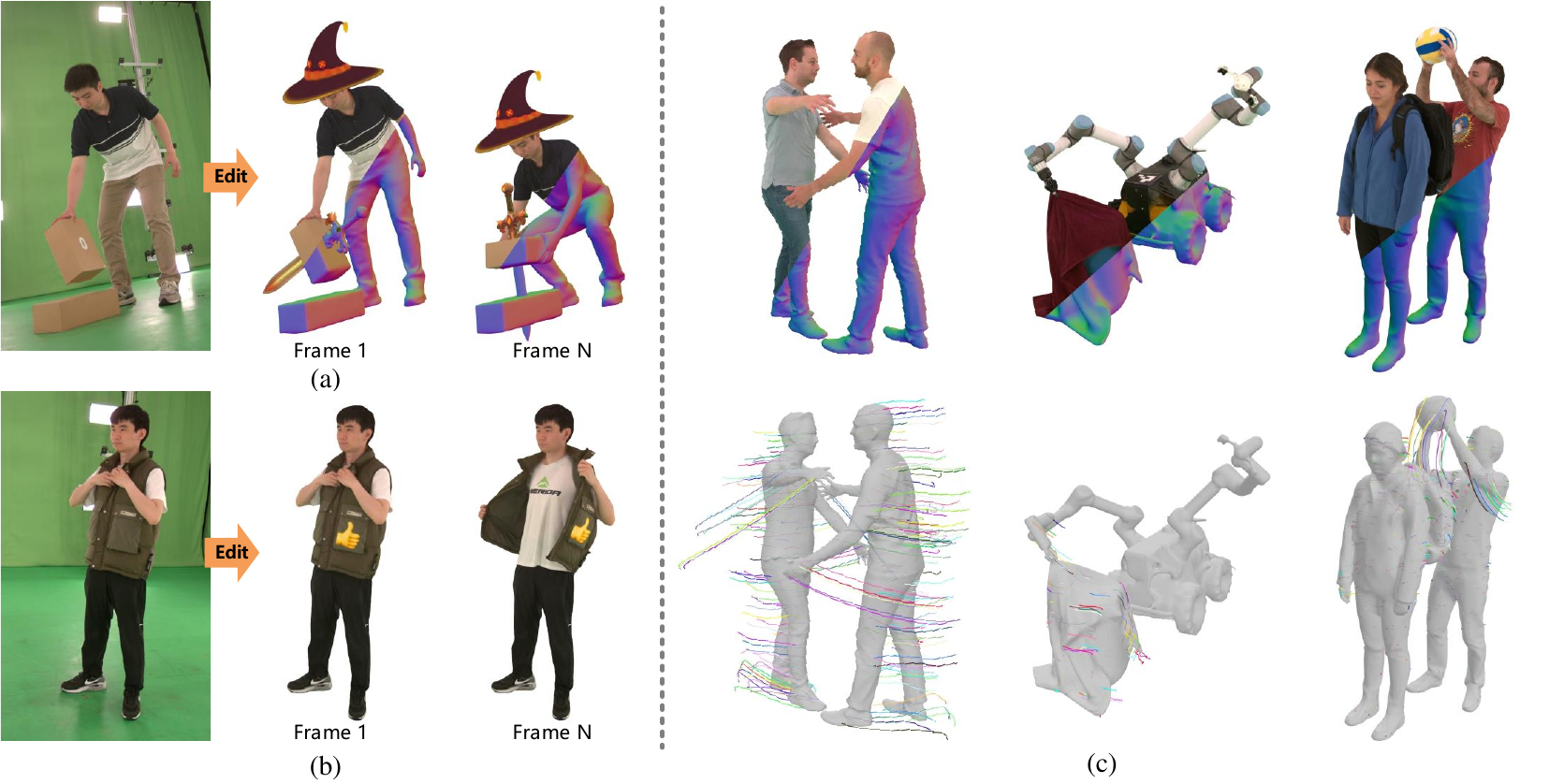}
\caption{Applications. 
(a) Object editing: virtual objects sync with dynamic surfaces. 
(b) Appearance editing: texture changes propagate across frames.
(c) General applicability to diverse scenes including multiple people, human-object interactions, and robotic motion.
}
\vspace{-0.5em}
\label{fig:editing}
\end{figure*}

\textbf{Tracking.}
We capture two sequences with AprilTags attached to the human body to evaluate tracking performance. Six AprilTags are placed on the front, back, arms, and legs, as illustrated in \cref{fig:tracking_comp}. For each tag, we track 5 key points: the center and 4 corner points. The 3D and 2D ground truth are provided by AprilTag detection. We report 3D and 2D Average Trajectory Error (ATE)~\cite{harley2022particle, zheng2023pointodyssey} in \cref{tab:comparison}, with visualizations of the tracking trajectories presented in \cref{fig:tracking_comp}. Our method achieves accurate tracking results, even on topology-changing surfaces (e.g., arms uncrossing).

\subsection{Ablations}

We evaluate the key contributions of our method, including Gaussian unbinding, surface re-meshing, and scene flow warping. For ablation, each component is disabled individually, with the results shown in \cref{tab:ablation} and in our supplementary materials.

\textbf{Gaussian unbinding}.
Since newly emerging surfaces cannot be modeled by fixed-topology meshes, our method unbinds Gaussians on topology-changing surfaces to ensure accurate reconstruction. Without unbinding, Gaussians remain attached to outdated faces, preventing updates to the surface topology and failing to address topology changes.

\textbf{Surface re-meshing}.
Following Gaussian unbinding, our method updates the underlying mesh to reconstruct new surfaces with the correct topology. Without this update, the geometry topology remains consistent with the initial input mesh, leading to suboptimal quality.

\textbf{Scene flow warping}.
For each frame initialization, we construct scene flow to warp the Gaussian Surfaces from the previous frame, effectively handling large deformations. This approach significantly enhances tracking quality and helps prevent the optimization process from getting trapped in local minima.


\section{Discussion}
\label{sec:conclusion}

\textbf{Applications.}
Our method enables continuous tracking throughout entire sequences. 
We present results for object editing and appearance editing in (a) and (b) of \cref{fig:editing} and in the supplementary video. 
For object editing, new objects can be inserted into dynamic scenes and move in sync with other surfaces through surface tracking. 
For appearance editing, modifications made in a single frame are propagated across frames via surface tracking. Our approach does not require a template and is applicable to various general dynamic scenes, including robots, multiple people, and interacting objects, as shown in \cref{fig:editing} (c).

\textbf{Limitations.}
\NAME may face challenges with complex or sudden topology changes, such as when a new person suddenly enters the scene. Its dependence on multi-view video data restricts its applicability in general public scenarios. Transparent and specular surfaces pose challenges for most surface reconstruction methods, particularly for Gaussian splitting. We leave handling such surfaces for future work.

\textbf{Conclusions.}
We propose \NAME, a unified method for high-quality appearance reconstruction, surface reconstruction, and 3D tracking. Our approach represents dynamic surfaces by binding Gaussians to mesh faces. For surfaces with changing topology, new surfaces are reconstructed by unbinding Gaussians. \NAME effectively handles a wide range of dynamic scenes, paving the way for new applications of Gaussian-based representations.

\section*{Acknowledgements}
This work was partially supported by the Swiss SERI Consolidation Grant "AI-PERCEIVE". The authors thank all the participants of the captured datasets. Part of the computations were performed on the ETH Zürich Euler cluster.

{
    \small
    \bibliographystyle{ieeenat_fullname}
    \bibliography{main}
}


\end{document}


\maketitle

In this supplementary document, we provide implementation details of \NAME and other baseline methods in~\cref{supp_sec:implement_details}. Then, we demonstrate more experiment results in~\cref{supp_sec:exp} with qualitative comparisons and ablation studies that further validate our method's effectiveness. Finally, we include an extensive discussion of ethics and societal impact of our approach in~\cref{supp_sec:ethics}.

In the supplementary video, we demonstrate \NAME's overall pipeline and its performance on various dynamic scenes. We also provide visual comparisons against baseline methods.

{
  \hypersetup{linkcolor=black}
  \tableofcontents
}

\noindent\rule{\linewidth}{0.4pt}

\section{Implementation Details}
\label{supp_sec:implement_details}

\subsection{\NAME Details}

\paragraph{Code releasing.} 
Our code is available at \todo{link}. We provide our implementation on publicly available datasets, allowing for reproducibility and further research. However, due to licensing restrictions, the data captured for \NAME testing will not be released.

\paragraph{Running Time.} 
Our method sequentially processes the video frames. Each frame requires approximately 5 minutes of training time on a single NVIDIA RTX 4090 GPU (running time varies depending on the face number). For frames where no topology changes are detected, we only perform fixed-topology surface reconstruction, reducing the processing time to approximately 2 minutes. Once training is complete, rendering is performed in real time, leveraging CUDA acceleration for core Gaussian splatting operations. A comparison of running times across different methods, using the same sequence, is provided in \cref{tab:comp_speed}. 

\begin{table}[ht]
\small
    \vspace{-0.5em}
    \centering
    \begin{tabular}{lcc}
        \toprule
        Method & Training (per frame) & Rendering\\
        \midrule
        HumanRF~\cite{icsik2023humanrf} & 4.9 min & 0.8 fps \\
        Dynamic 3D GS~\cite{luiten2024dynamic} & 1.5 min & 203 fps \\
        PhysAvatar~\cite{zheng2024physavatar} & 1.1 min & 218 fps \\
        2DGS~\cite{huang20242d} & 7.1 min & 231 fps \\
        \NAME (Ours) & 4.6 (or 2.1) min & 188 fps \\
        \bottomrule
    \end{tabular}
    \vspace{-0.5em}
    \caption{Runtime comparisons on a single RTX 4090. \NAME requires 2.1 min when no topology changes are detected.}
    \vspace{-0.5em}
\label{tab:comp_speed}
\end{table}

\paragraph{Initial Input for the First Frame.} \NAME requires a mesh as input for each frame. For the first frame, we employ an RGB-D based multi-view reconstruction method \cite{collet2015high} to generate the initial mesh. The reconstructed mesh is then down-sampled to contain between 100,000 and 200,000 faces, with the exact count varying according to scene complexity. We attach $N = 6$ Gaussians to each face, resulting in approximately 600,000 to 1,200,000 total Gaussians (note that the number of faces may dynamically change due to topology updates). The Gaussian appearance parameters for the first frame are initialized using the mesh texture, while the initial opacity, scale, and rotation parameters are set to predefined values.

\paragraph{Remeshing Details.} 
As introduced in Sec. 3.5 of the main paper, after generating new faces, we update the underlying mesh topology by integrating newly generated faces with the original mesh.
This process involves removing specific faces from the original mesh, identifying corresponding newly generated faces, and seamlessly connecting them. 

This method begins by identifying original faces where the unbound weight exceeds a predefined threshold. 
These faces are then grouped by their connected components. We delete any connected components that contain more faces than a specified threshold. Next, we create a voxel volume to record the positions of unbound Gaussians from deleted faces. Within this volume, we identify newly generated faces and remove isolated faces based on their connected components, preparing them for integration with the remaining original mesh. The connection process involves two steps of vertex matching: first, for each vertex $x$ on the boundary of newly generated faces, we locate its closest vertex $y$ on the original mesh boundary, set their positions to $y$, and merge them; then, for unmatched vertices on the original mesh boundary, we find their closest vertices on the new face boundary and perform similar alignment and merging operations. Finally, we complete the mesh reconstruction through edge flipping and hole filling operations to ensure a seamless surface.

\subsection{Baseline Details}

\paragraph{HumanRF~\cite{icsik2023humanrf}.} 
As the official mesh extraction code for HumanRF is not publicly available, we implemented our own version following their paper. While we use Marching Cubes for mesh extraction, the raw outputs often contain undesirable internal surfaces, such as those inside the human body. To address this issue, we additionally generate outer surfaces using TSDF fusion, then remove any mesh faces that are far from these TSDF-extracted surfaces. We also implement light annotations in HumanRF to reduce light bloom artifacts. We capture a background frame to detect intense light sources and mask the affected image regions. While this enhancement improves the overall quality, it does not entirely eliminate the artifacts.

\paragraph{Dynamic3DGS~\cite{luiten2024dynamic}.} 
The original Dynamic3DGS paper primarily focuses on rendering quality rather than geometric reconstruction. As it does not provide a dedicated surface reconstruction method, we employ TSDF fusion techniques similar to our surface generation approach, combining depth images from multiple views to obtain the final mesh. Due to the inherent limitations of Gaussian splatting, the resulting reconstruction exhibits considerable noise in the geometry.

\paragraph{2D Gaussian Splatting~\cite{huang20242d}.} 
2DGS is designed for reconstructing static scenes, and we use it to process each frame independently. For each frame, we initialize the point cloud using the refined point cloud from \cite{collet2015high}. These point clouds are the ones used for rendering depth inputs from IR cameras. We down-sample each frame's point cloud to 600,000 points before processing. To enhance geometry consistency, we incorporate a mask loss similar to our formulation in Eq. (6) during the training stage. Unlike \NAME and other baselines that leverage temporal information across frames, 2DGS reconstructs each frame independently. As a result, 2DGS is less robust and more prone to overfitting in our 47-view setting, leading to floater artifacts and notable temporal jittering. A similar overfitting trend is also observed in another experiment, where reducing input views from 200 to 47 on the Mip-NeRF dataset causes the training PSNR to increase (29.2 to 31.5), while the test PSNR declines (28.7 to 26.9).

\paragraph{PhysAvatar~\cite{zheng2024physavatar}.} 
While PhysAvatar's original paper describes a pipeline beginning with mesh tracking followed by clothing reconstruction and simulation, it does not explicitly mention SMPL-X dependency. However, their released implementation utilizes SMPL-X for improved tracking robustness, particularly in hand regions, as confirmed by the authors. Without SMPL-X initialization, their method relies on inertial estimates for full-body initialization between frames, similar to Dynamic3DGS.

For a fair comparison across our diverse sequences containing single humans, multiple humans, and non-human objects, we implement two variants of PhysAvatar: one using inertial initialization and another additionally using SMPL-X deformation for human vertices. As the SMPL-X fitting code was not publicly available at the time of submission, we adopted the approach from X-Avatar~\cite{shen2023xavatar} to fit SMPL-X using multi-view images and reconstructed meshes. This fitting is done with a multi-stage pipeline: first extracting 2D keypoints using OpenPose and triangulating them to 3D with specific filtering for unstable hand predictions. The SMPL-X parameter optimization then proceeds through three stages: we first initialize the parameters using the filtered 3D keypoints, then refine body pose and shape parameters using the scan geometry, and finally optimize hand poses and facial expressions using 3D landmarks.
\section{Additional Experiments}
\label{supp_sec:exp}

\subsection{Comparisons with Baselines}

\begin{figure*}[ht]
\includegraphics[width=\linewidth, trim=0 0 0 0,clip]{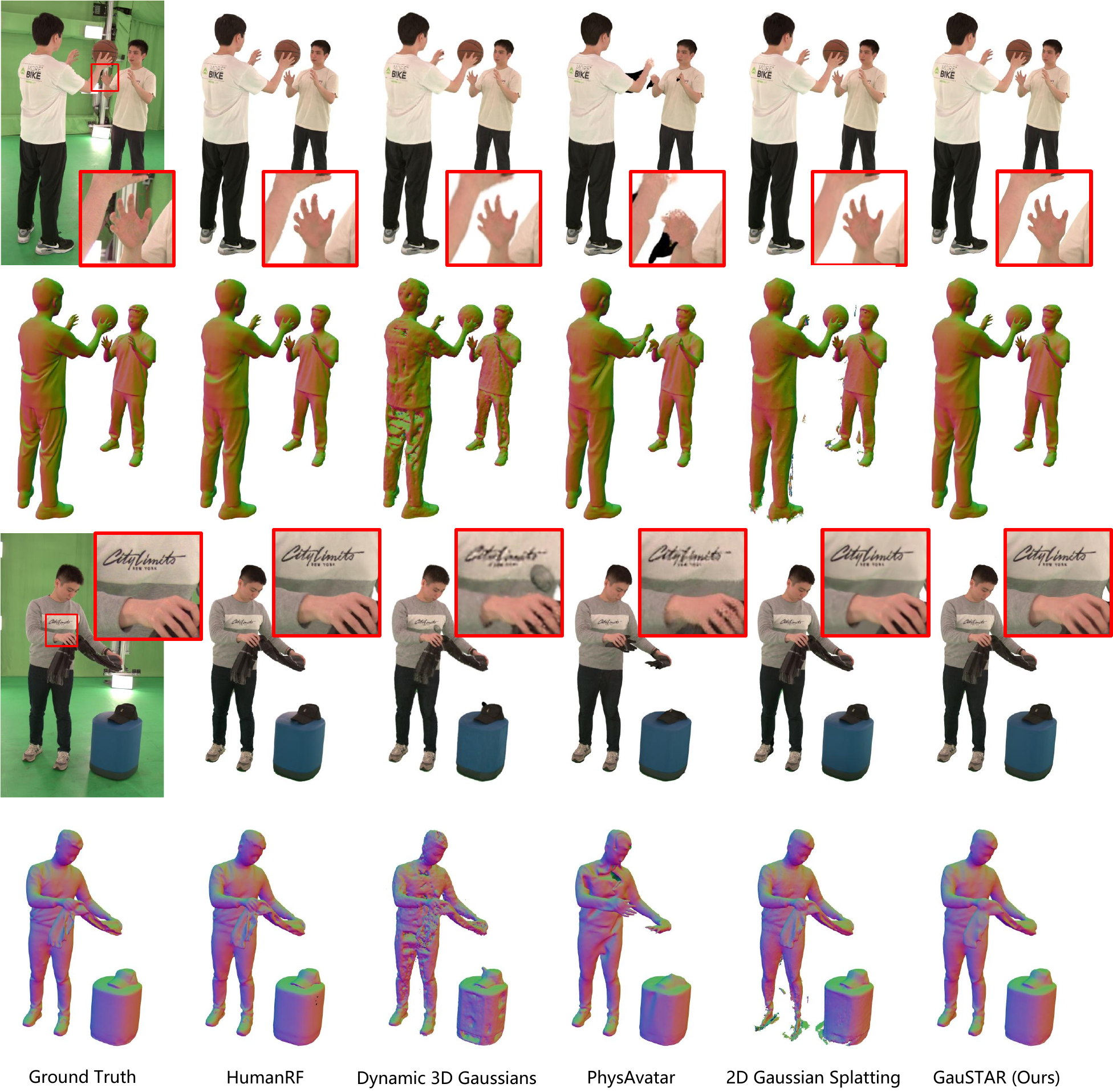}
\caption{Comparisons of appearance and geometry reconstruction. HumanRF offers overall good visual quality but lacks tracking capabilities. Dynamic 3D Gaussians produces blurry renderings and noisy surfaces. PhysAvatar struggles with handling topology changes, while 2D Gaussian Splatting faces challenges with both tracking and floating artifacts. In contrast, \NAME delivers high-quality reconstruction and effectively manages topology changes.}
\label{fig:comp_supp}
\end{figure*}

\begin{figure*}[ht]
\includegraphics[width=\linewidth, trim=0 0 0 0,clip]{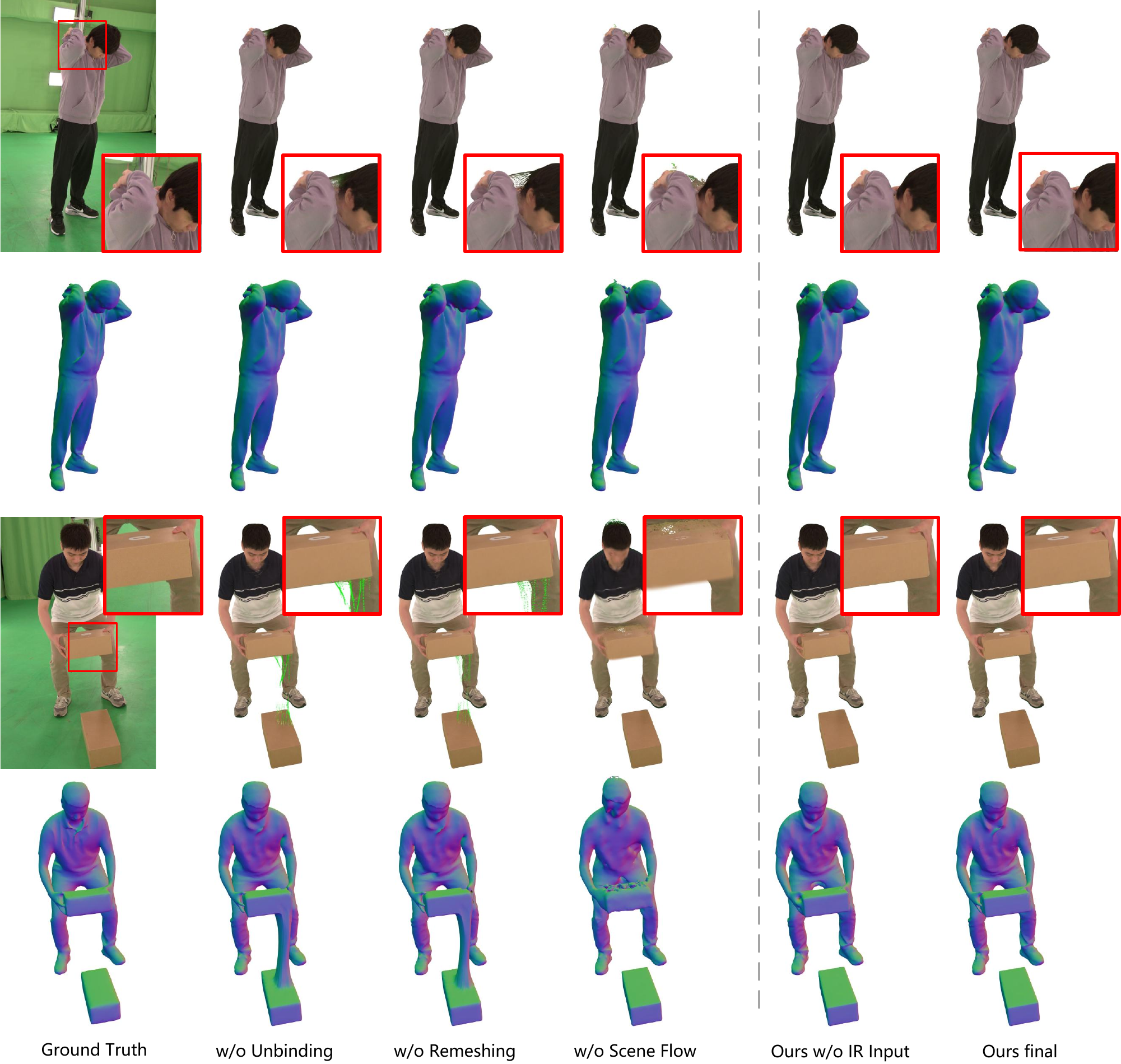}
\caption{Qualitative results for ablation study. Unbinding and re-meshing are crucial for handling topology changes, and scene flow ensures robust tracking of large movements. Our method without IR input yields a similar quality to the full version of our method but needs additional data prepossessing.}
\label{fig:supp_ablation}
\end{figure*}

We provide additional qualitative comparisons with HumanRF~\cite{icsik2023humanrf}, Dynamic 3D Gaussians~\cite{luiten2024dynamic}, PhysAvatar~\cite{zheng2024physavatar}, and 2D Gaussian Splatting~\cite{huang20242d} in~\cref{fig:comp_supp}.

HumanRF trains each video segment independently. This approach leads to slow rendering times and inconsistent tracking. Due to its independent segment training, HumanRF struggles with strong occlusions where most cameras cannot observe certain regions (Fig. 4 in the main paper). In contrast, our method demonstrates greater robustness to occlusion through consistent tracking and scene flow warping.

2DGS is a static scene reconstruction method, and we use it to process frames independently for reconstructing dynamic surfaces. However, its reconstruction quality varies significantly across frames, with some frames showing impressive results while others exhibit notable artifacts. Furthermore, as a single-frame method, it produces temporally unstable reconstructions and does not provide tracking capabilities.

Dynamic3DGS employs inertial estimation to initialize Gaussians for subsequent frames. While it supports tracking, the geometric quality is limited. Moreover, without an underlying mesh constraint, Gaussians can move freely in space, resulting in inconsistent tracking.

PhysAvatar maintains consistent mesh tracking and achieves high-quality reconstruction for clothed humans under normal motions.  However, its fixed-topology assumption fundamentally limits its ability to handle dynamic scenes where topology changes occur. In such cases, it fails dramatically when encountering topology changes due to its inability to handle such modifications. 

\subsection{Qualitative Results for Ablations}

We illustrate the impact of our key components through qualitative comparisons in \cref{fig:supp_ablation}. Without the unbinding and remeshing components, our method fails to properly handle topology changes, resulting in incorrectly merged geometries between the human head and hood in the first two rows, and erroneous connections between the separate boxes. The scene flow warping initialization proves crucial as well; without it, Gaussians become trapped in local minima and cannot properly redistribute, leading to significant geometric artifacts.

\subsection{Additional Scene Flow Ablation}

\begin{figure}[t]
\includegraphics[width=\linewidth, trim=10 0 0 0,clip]{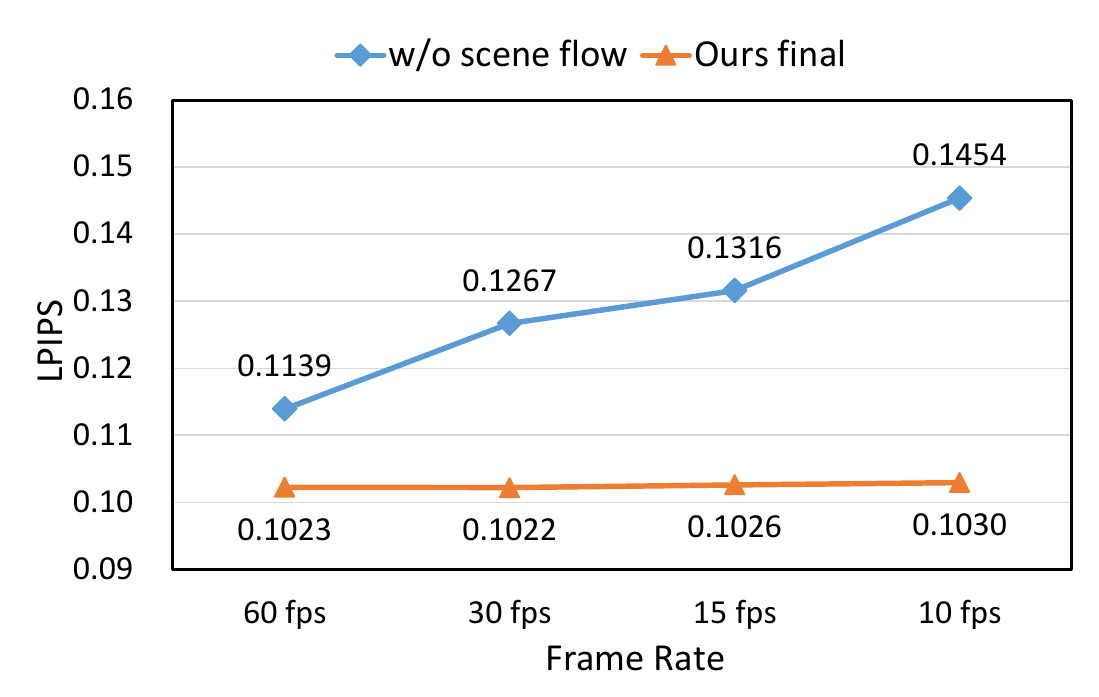}
\caption{Scene flow warping ablation. We capture a sequence at 60 FPS and down-sample it to 30 FPS, 15 FPS, and 10 FPS, increasing motion between frames. Our method consistently performs well across frame rates, while the method without scene flow shows higher errors as the motion between frames increases.}
\label{fig:flow_ablation}
\end{figure}

To demonstrate the effectiveness of our scene flow warping in handling large motions, we evaluate reconstruction quality under varying degrees of inter-frame movement. We conduct this experiment by capturing a sequence at 60 FPS and systematically down-sampling it to 30 FPS, 15 FPS, and 10 FPS, effectively increasing the magnitude of motion between consecutive frames. Comparing our full method against a variant without scene flow warping reveals that our approach maintains consistent reconstruction quality across all frame rates, while the ablated version shows progressively deteriorating performance as inter-frame motion increases.

\subsection{\NAME with RGB input}
Our method does not necessarily require the IR depth input. For example, we can use multiview stereo to create a rough depth map. Here we use rendered depth from HumanRF for its robustness and smoothness, which has fewer artifacts. As shown in the last two columns of \cref{fig:supp_ablation}, \NAME works almost equally well compared to the version with depth input. This means \NAME can be utilized for capture setup with RGB input only.

\begin{figure}[t]
    \includegraphics[width=\linewidth, trim=0 0 0 0,clip]{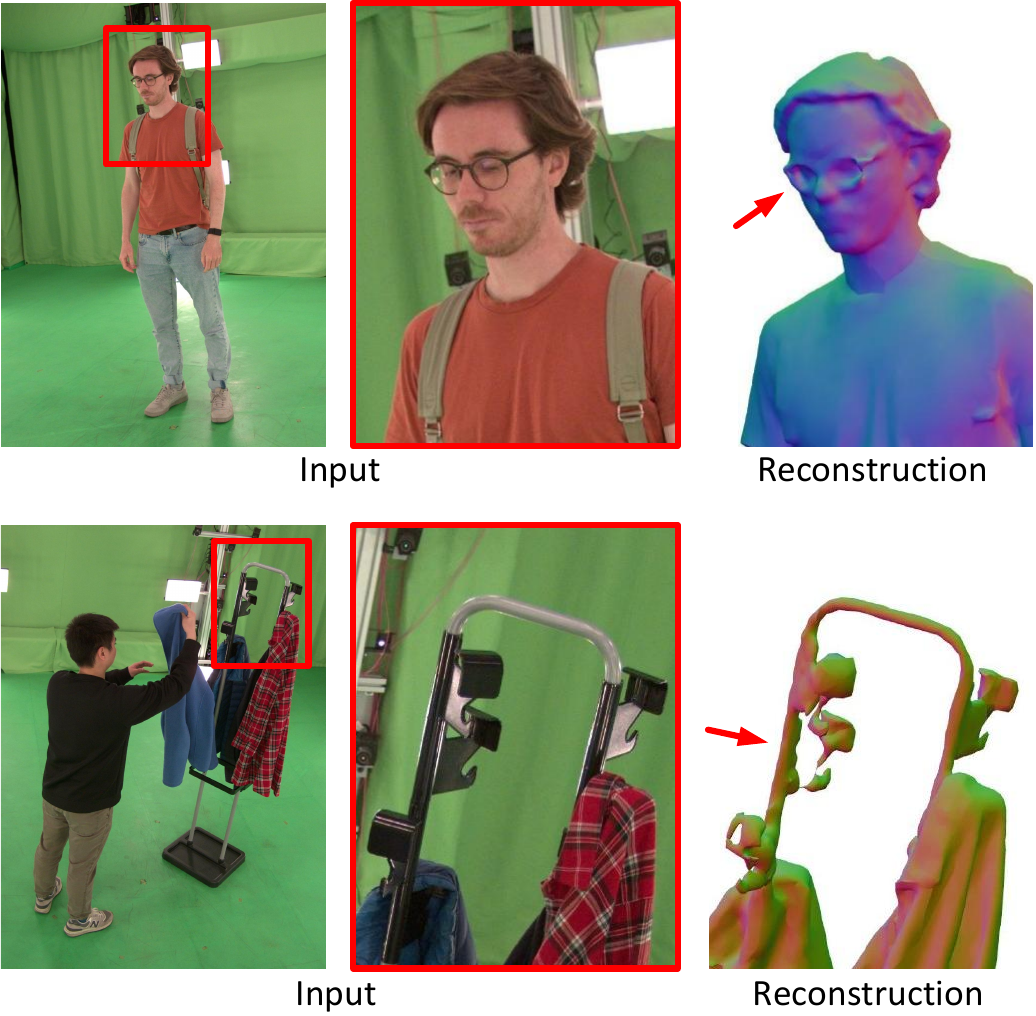}
    \vspace{-1.5em}
    \caption{
    Failure cases on transparent and specular surfaces.
    }
    \vspace{-0.5em}
\label{fig:failure_case}
\end{figure}

\subsection{Failure Cases}
We present failure cases in \cref{fig:failure_case}, highlighting the challenges faced by our method. In particular, transparent and specular surfaces remain difficult for most surface reconstruction techniques, including Gaussian splitting, due to their complex light interactions and lack of reliable depth cues. Accurately reconstructing such surfaces requires more advanced strategies, which we leave as future works.
\section{Ethics and Societal Impact Discussion}
\label{supp_sec:ethics}

Our data collection procedure has been reviewed and approved by the responsible Institutional Review Board.
All subjects voluntarily participated in the data collection process and were fully informed about the intended use of the data in research.

\NAME enables the digitization of general dynamic scenes from multi-view captures, which has broad applications in visual effects, robotics, and virtual production. As our method can reconstruct and track detailed surface changes, there are potential concerns about privacy and surveillance when applied to scenes involving human activities. Such concerns must be addressed before deploying this technology in commercial products. Our goal with this work is to enable beneficial applications such as human-robot interaction, markerless motion analysis, and immersive telepresence. Our system represents a technical advancement in computer vision that can benefit numerous fields from industrial automation to cultural preservation. While we cannot prevent potential misuse of such technology, we believe in transparent research practices, including detailed technical discussions and code release. This openness allows the research community to better understand both the capabilities and limitations of such systems, and to develop appropriate safeguards against concerning applications.


{
    \small
    \bibliographystyle{ieeenat_fullname}
    \bibliography{main}
}
